\documentclass{article} 
\usepackage{iclr2024_conference,times}
\usepackage{graphicx}

\usepackage{amsmath,amsfonts,bm}

\def\eqref#1{equation~\ref{#1}}

\def\1{\bm{1}}

\DeclareMathAlphabet{\mathsfit}{\encodingdefault}{\sfdefault}{m}{sl}
\SetMathAlphabet{\mathsfit}{bold}{\encodingdefault}{\sfdefault}{bx}{n}

\usepackage{hyperref}
\usepackage{url}

\title{How Consistent are Clinicians? \\ Evaluating the Predictability of Sepsis \\ Disease Progression with Dynamics Models}

\author{Unnseo Park, Venkatesh Sivaraman \& Adam Perer \\ 
Human-Computer Interaction Institute\\
Carnegie Mellon University\\
Pittsburgh, PA 15213, USA \\
\texttt{upark@andrew.cmu.edu} \\
\texttt{\{venkats,adamperer\}@cmu.edu}
}

\iclrfinalcopy 
\begin{document}

\maketitle

\begin{abstract}
Reinforcement learning (RL) is a promising approach to generate treatment policies for sepsis patients in intensive care. 
While retrospective evaluation metrics show decreased mortality when these policies are followed, studies with clinicians suggest their recommendations are often spurious. 
We propose that these shortcomings may be due to lack of diversity in observed actions and outcomes in the training data, and we construct experiments to investigate the feasibility of predicting sepsis disease severity changes due to clinician actions. 
Preliminary results suggest incorporating action information does not significantly improve model performance, indicating that clinician actions may not be sufficiently variable to yield measurable effects on disease progression. 
We discuss the implications of these findings for optimizing sepsis treatment. 
\end{abstract}

\section{Introduction}

Sepsis is a leading cause of death in hospitals, and there is currently little clinical consensus around best practices for treatment~\citep{CentersforDiseaseControlandPrevention2021}. Several recent works have applied reinforcement learning (RL) methods in efforts to support clinicians' decision-making on sepsis patients in the intensive care unit (ICU). While these algorithms have shown promise when evaluated using off-policy policy evaluation (OPE) methods, they have also been critiqued for recommending incorrect and even dangerous treatment plans, particularly for more severely ill patients~\citep{Jeter2019,Sivaraman2023}. Due to ethical concerns around prospectively evaluating these models, it is currently an open question whether it is possible to derive policies from public observational datasets that truly improve current clinical practice.

\begin{figure}[h]
\begin{center}
    \includegraphics[width=\textwidth]{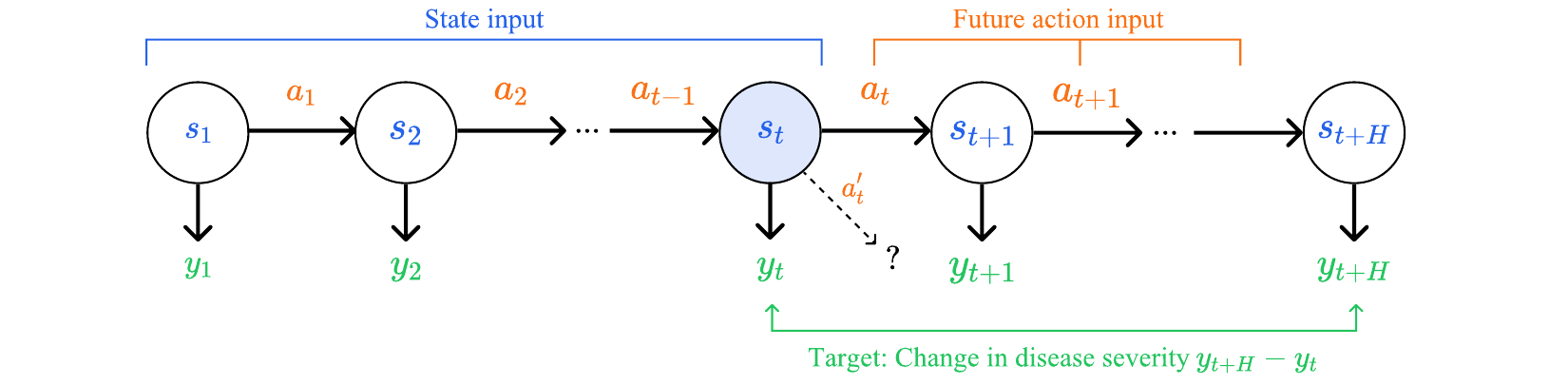}
    \end{center}
    \caption{Markov decision process model for patients with sepsis in the ICU. $s_t$ represents the patient state at time $t$, $a_t$ represents a treatment action, and $y_t$ represents a function of the state that captures the patient's disease severity. Brackets indicate how these values are used in our experiment.}
    \label{fig:trajectory-schematic}
\end{figure}

To produce meaningful recommendations with adequate data support, we propose that patient trajectory datasets should exhibit \textit{diversity in observed actions} that correlates with differences in outcomes conditioned on a particular state. In the RL formulation shown in Fig. \ref{fig:trajectory-schematic}, we assume that for a given state $s_t$ we can estimate not only the cumulative reward of taking observed action $a_t$, but also the reward for taking a different action $a^{\prime}_t$. This would allow the offline-trained RL agent to accurately choose between $a_t$ or $a^{\prime}_t$ despite having only observed directly the results of the former action.

It is difficult to measure action and outcome diversity conditioned on states directly, since this can depend heavily on how the state is represented. Instead, we constructed an experiment in which we trained transformer-based \textit{dynamics models} to predict future disease severity given a patient's state and optionally the treatment actions that were taken over the subsequent hours. If clinician actions are diverse and have an effect on outcomes, then the action information should improve a model's ability to predict future observed disease severity. Below we present preliminary results from this experiment and discuss their implications for future efforts to optimize sepsis treatment.

\section{Related Work}
Several studies~\citep{Raghu2017,Peng2018,Yu2019,liu_offline_2021,ju_reduce_2021} have applied RL to train treatment policies for sepsis patients.
For instance, the AI Clinician agent proposed by \cite{Komorowski2018} utilized tabular $Q$-learning, while subsequent works have proposed combining deep RL with kernel-based RL~\citep{Peng2018}, applying deep inverse RL~\citep{Yu2019}, integrating physiological models~\citep{nanayakkara_unifying_2022}, and improving sample efficiency by focusing on important timesteps~\citep{liang_treatment_2022,ju_reduce_2021}). \cite{tang_leveraging_2023} further introduced an approach that leverages factored action spaces to improve the efficiency of offline RL in healthcare settings. 

Other research highlights the challenges and shortcomings of these algorithmic approaches. For example, \cite{Killian2020} found that even powerful sequential models were unable to accurately separate patient state representation by mortality. \cite{Jeter2019} suggested that the $Q$-learning approach may learn to recommend dangerous treatments for severely ill patients because the most common clinician actions have consistently low rewards, whereas rarer actions may have high but noisy estimated values. \cite{Gottesman2020} proposed improving RL policy evaluation by identifying timepoints with a high OPE weight, while \cite{Ji2021} selected and visualized trajectories that may explain policy behavior. To our knowledge, however, predicting future disease outcomes from actions has not been examined as a way to evaluate the feasibility of off-policy RL in sepsis. 

\section{Methods}
\subsection{Data and Preprocessing}
Patient trajectory data was extracted following \cite{Komorowski2018} and \cite{Killian2020} from MIMIC-IV~\citep{johnson2020mimic} and the eICU Collaborative Research Database ~\citep{pollard2018eicu}.\footnote{While previous work has generally used MIMIC-III, the AI Clinician modeling procedure has been shown to yield consistent results in the two versions of MIMIC~\citep{Sivaraman2023}.}\footnote{Preprocessing and modeling code available at \url{https://github.com/cmudig/AI-Clinician-MIMICIV}.} Data was aggregated at one-hour intervals, and patients with more than 14 days in the ICU were excluded. Missing data was imputed using a transformer-based autoencoder model. This resulted in a total of 2,060,446 timesteps from 33,779 patients. 


The state space for our models consisted of 60 normalized observation variables (vitals, labs, prior treatments, and fluid balances) and 35 demographic variables (age, gender, and Elixhauser comorbidities). The action space comprised log-transformed continuous-valued dosages of IV fluids and vasopressors. Three widely-used severity metrics were used as outcomes: the Sequential Organ Failure Assessment (SOFA) score, the Systemic Inflammatory Response Syndrome (SIRS) score, and Shock Index. Actions and disease severity were $z$-transformed for model input and output.

\subsection{Models}

\paragraph{Dynamics models} Our experiment utilized decoder-only transformer models, where each input ``token'' comprised embeddings of the patient's observed state, demographics, and actions.\footnote{We conducted the same experiments with linear and recurrent networks as well as XGBoost models, but found that transformers yielded the best performance.} The model consisted of two transformer blocks, each comprising 4 self-attention layers, each with 16 attention heads and a total dimension of 1024. The first transformer block took the state and demographic embeddings as input, while the second transformer block added embedded clinician actions. Models were trained on the future disease severity task along with three other proxy tasks: (1) predicting the current state of the patient, (2) predicting whether the current state is the last step in the patient's trajectory, and (3) predicting whether two embeddings correspond to states that are adjacent in time. The proxy tasks were included only to improve the model's convergence and generalizability, and results for these tasks are not shown. 


\paragraph{Behavior cloning}
While the dynamics models described above aimed to predict the difference in disease severity as a function of states and actions, we also trained behavior cloning models to predict clinician actions as a function of states. These models utilized the first transformer block from above to encode the state observations and demographics, then applied a two-layer feedforward network to simultaneously predict fluid and vasopressor dosages at one-hour intervals up to 6 hours ahead. 

\section{Experiment Results}
\subsection{Influence of Action Inputs on Disease Severity Predictions} \label{sec:disease-severity-results}

\begin{figure}[h]
\begin{center}
\includegraphics[width=\textwidth]{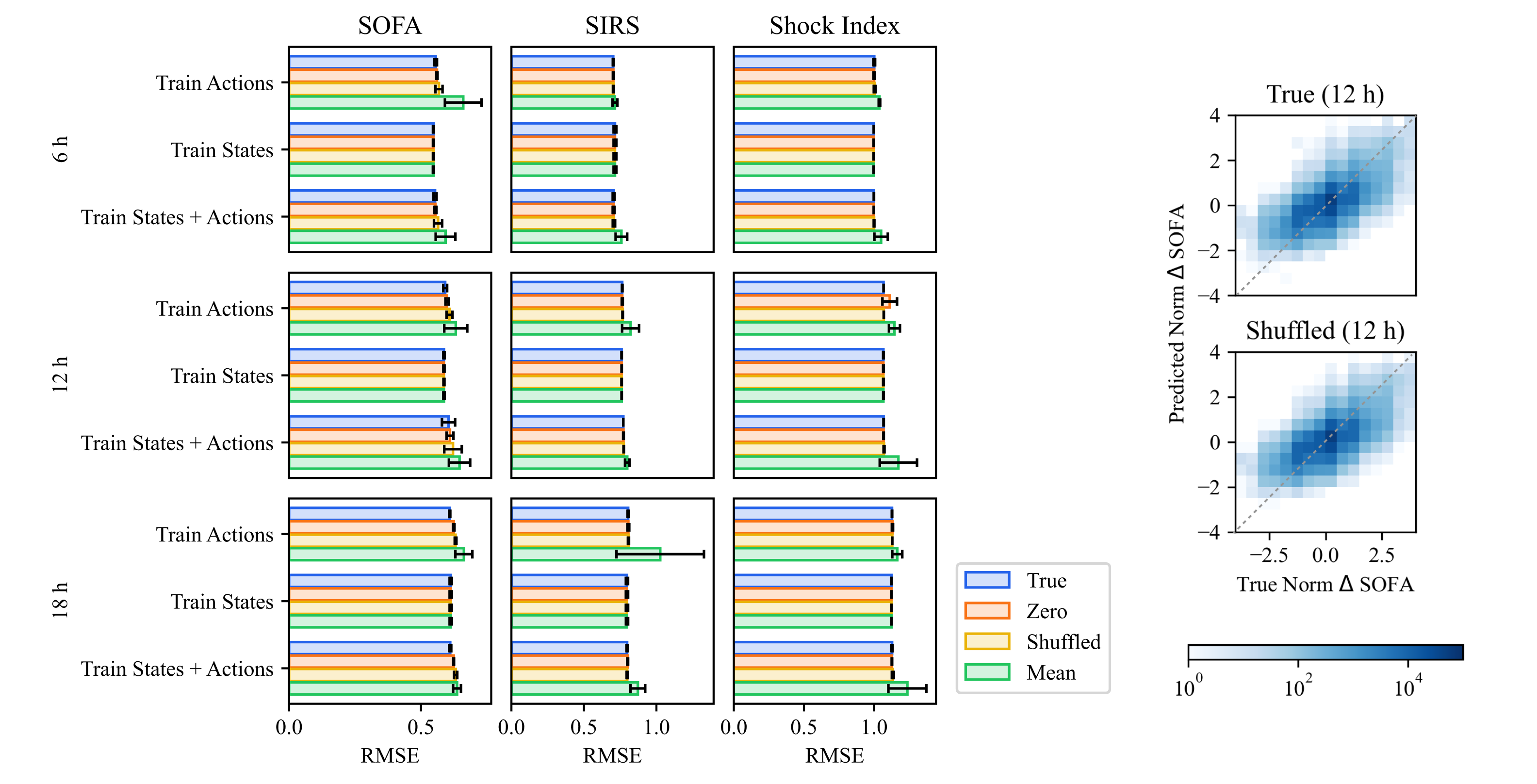}
\end{center}
\caption{Left: RMSE (lower is better) of the predicted change in disease severity across training schemes (``Train Actions'', ``Train States,'' and ``Train States + Actions'') and action inputs at test time (\textbf{True}, \textbf{Zero}, \textbf{Shuffled}, and \textbf{Mean}). Error bars indicate the standard deviation across three random weight initializations. Note that all units are in $z$-scaled space, so an RMSE of 1 corresponds to 1 standard deviation in the severity metric. Right: example histograms comparing true and predicted changes in SOFA score at 12 hours ahead, in the \textbf{True} and \textbf{Shuffled} evaluation conditions.}
\label{fig:transformer-results}
\end{figure}

Three groups of models, totaling 81 dynamics models, were trained to predict changes in future disease severity. The first group was trained with both future action information and state information. In contrast, the second group, featuring identical architectures, had all future actions set to the mean action values (effectively removing them from training). The last group also shared identical architectures, but was trained without the information about states. Furthermore, disease severity changes were measured according to the three metrics described above at 6 hours, 12 hours, and 18 hours ahead. Each model configuration was trained and evaluated across three random weight initializations. We then conducted four evaluations for each model by generating predictions on variants of the test dataset: \textbf{True} (original treatment actions), \textbf{Zero} (all dosage values set to zero), \textbf{Shuffled} (real but randomly-permuted dosages), and \textbf{Mean} (all actions replaced with the mean dosages). Fig. \ref{fig:transformer-results} shows the root mean squared error (RMSE) of these predictions in $z$-scaled space, as well as two examples comparing model predictions to ground-truth.

Overall, the RMSE was almost constant across training conditions and action input types except for the \textbf{Mean} condition, which generally showed higher error and variance across initializations when actions were used in training (likely because consistently receiving nonzero fluids and vasopressors is a highly unusual input). Among the other three conditions, the range of RMSEs was within 0.05 for SIRS and Shock Index, and within 0.1 for SOFA. Furthermore, performance in the \textbf{True} condition was highly similar whether or not actions were provided during training. This null result suggests that actions did not substantially improve the model fit, consistent with our hypothesis that they are not diverse enough for policy learning. In addition, models trained without state information showed similar trends, indicating that action information is largely redundant with the states.

\subsection{Prediction of Future Actions with Behavior Cloning} \label{sec:action-results}

To evaluate the predictability of actions from states more directly, we trained 3 replicates of the behavior cloning model with different random weight initializations. If these models showed a strong fit to the data, one could infer that actions were fully consistent and predictable across clinicians. 

\begin{figure}[h]
\begin{center}
\includegraphics[width=\textwidth]{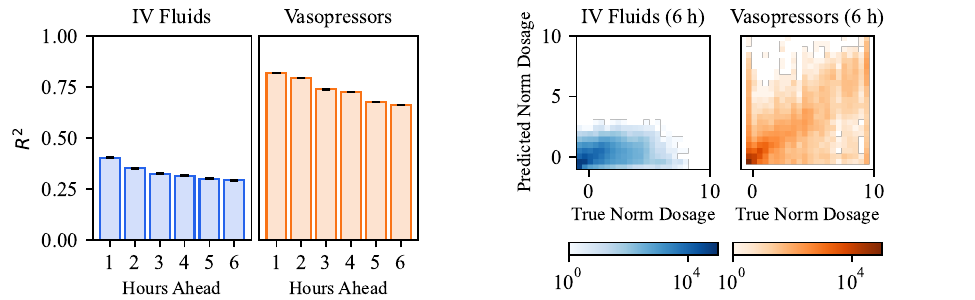}
\end{center}
\caption{Left: correlations between true and predicted normalized actions from 1 to 6 hours ahead. Right: example histograms of correlations between true and predicted normalized actions at 6 hours.}
\label{fig:action-predictions}
\end{figure}

Fig. \ref{fig:action-predictions} shows that the average $R^2$ correlations between the true and predicted actions (in log-transformed and $z$-scaled units) were overall low, particularly after several hours. IV fluid predictions were markedly less correlated with the true values than vasopressor predictions, perhaps because (1) vasopressors are more commonly zero than fluids, increasing the overall predictability of vasopressor use, or because (2) the amount of IV fluid used is generally more clinician-dependent. The regression models also appeared to struggle with the wide range of fluid dosage values, and tended to predict values within a more constrained range (Fig. \ref{fig:action-predictions}, third panel). 

Aside from the possible modeling issues in the IV fluid predictions, the low correlations across both treatments suggest there is in fact some diversity in clinician actions that could benefit policy learning. However, action diversity does not necessarily correspond to observable differences in outcomes, since there is likely a range of treatment dosages that correspond to similar effects for a given patient state. The results in the preceding section suggest that even when dosage differences exist, they may not yield sufficient differences in outcomes to provide a useful signal to an RL agent.

\section{Discussion}

This work explored the impact of clinician actions on the predictability of future changes in sepsis disease severity, in order to gain insight into whether actions have sufficient diversity to support accurate RL-based policies. We found that action information does \textit{not} confer substantive improvements in dynamics model fit, as our transformer models could predict future disease severity almost equally well with or without true actions as input. Taken alone, the dynamics model results in Sec. \ref{sec:disease-severity-results} might suggest that actions are fully predictable from the states and there was no need to learn from the action inputs. This echoes results from~\cite{beaulieu-jones_machine_2021}, who critique patient risk predictions as ``looking over the shoulders of clinicians.'' However, action prediction (Sec. \ref{sec:action-results}) was still fairly noisy, indicating that while variation in actions exists, it is not enough to cause measurable differences in outcomes in our sepsis cohort. Rather, the outcome differences we observe may be more driven by unobserved patient variables or natural random variation.

Some of the observed lack of diversity in actions on MIMIC data may be due to inherent challenges in working with patient trajectories. For instance, there may only be a small number of treatment possibilities that are clinically feasible and safe, limiting the space of actions that clinicians could take. Clinicians may also tend to choose actions in predefined patterns, such as monotonically increasing or decreasing dosages, that appear diverse yet lead to consistent outcomes. Alternatively, missing data imputation could have caused patient states and actions to appear more consistent than they really are. These obstacles are likely to exist in any patient treatment dataset, underscoring the importance of using learning methods that are robust to missingness and a constrained action space.

Another possible explanation for our results is that our models simply didn't learn to use actions effectively, and a better model formulation might yield more pronounced differences between the ``Train States'' and ``Train States + Actions'' models. It is impossible to determine \textit{a priori} whether there exists a more effective way to use actions, but we conjecture that if such a method exists, it would likely require more clinically-informed descriptions of actions than what has currently been explored in the literature. For instance, models could use other treatments such as antibiotics and mechanical ventilation, contextualize actions using the patient's physiological state, or limit the training data to only the most important decision points. Future work should incorporate clinician guidance on how to meaningfully encode treatments to further test the effects of action information.

This work highlights the importance of diversity in data sources when building medical recommendation models. While it has been extremely valuable in developing and exploring ways to improve sepsis treatment, the MIMIC dataset is sourced from a single well-resourced hospital in Boston~\citep{johnson2020mimic}, where clinicians are likely to be consistent and compliant with existing practice guidelines. Human-centered ML efforts undertaken in collaboration with clinicians and medical data experts can also inspire more clinically-relevant and performant model formulations, such as focusing on the emergency department (a higher-stress environment that is less specialized towards sepsis than the ICU) or building smaller models that are relevant to specific subgroups of patients~\citep{Sivaraman2023}. Through these research directions, applied ML efforts may be able to better utilize available observational data to improve sepsis treatment recommendation.

\subsubsection*{Acknowledgments}
The authors would like to thank Dr. Jeremy Kahn, Andrew King, Jason Kennedy, and the anonymous reviewers for feedback on the manuscript. This work was supported by a National Science Foundation Graduate Research Fellowship (DGE2140739), and by the Carnegie Mellon University Center of Machine Learning and Health.

\bibliography{main}
\bibliographystyle{iclr2024_conference}


\end{document}